\documentclass[letterpaper, 10 pt, conference]{ieeeconf}  

\IEEEoverridecommandlockouts                              

\title{\LARGE \bf
Utilising Explainable Techniques for Quality Prediction in a Complex Textiles Manufacturing Use Case
}

\author{Briony Forsberg$^{1}$, Henry Williams$^{1}$, Bruce MacDonald$^{1}$, Tracy Chen$^{2}$, Reza Hamzeh$^{2}$, Kirstine Hulse$^{2}$
\thanks{*This work was funded by Bremworth Carpets and Rugs, New Zealand.}
\thanks{$^{1}$Centre for Automation and Robotic Engineering Science, The University of Auckland, New Zealand
        {\tt\small bfor519@aucklanduni.ac.nz}}%
\thanks{$^{2}$Bremworth Ltd, Auckland, New Zealand}%
}

\usepackage[table,xcdraw]{xcolor}

\usepackage{xcolor, soul}
\sethlcolor{yellow}

\usepackage{graphicx}

\usepackage[english]{babel}

\begin{document}

\maketitle
\thispagestyle{empty}
\pagestyle{empty}


\begin{abstract}

This paper develops an approach to classify instances of product failure in a complex textiles manufacturing dataset using explainable techniques. The dataset used in this study was obtained from a New Zealand manufacturer of woollen carpets and rugs. In investigating the trade-off between accuracy and explainability, three different tree-based classification algorithms were evaluated: a Decision Tree and two ensemble methods, Random Forest and XGBoost. Additionally, three feature selection methods were also evaluated: the SelectKBest method, using chi-squared as the scoring function, the Pearson Correlation Coefficient, and the Boruta algorithm. Not surprisingly, the ensemble methods typically produced better results than the Decision Tree model. The Random Forest model yielded the best results overall when combined with the Boruta feature selection technique. Finally, a tree ensemble explaining technique was used to extract rule lists to capture necessary and sufficient conditions for classification by a trained model that could be easily interpreted by a human. Notably, several features that were in the extracted rule lists were statistical features and calculated features that were added to the original dataset. This demonstrates the influence that bringing in additional information during the data preprocessing stages can have on the ultimate model performance. 

\end{abstract}


\section{Introduction}

There are many uncertainties within manufacturing processes that can affect the quality of the final product \cite{112-ref1}. Root Cause Analysis (RCA) seeks to prevent a problem from recurring by understanding the causal mechanism behind a change from a desirable state to an undesirable one \cite{112-ref4}. RCA enables manufacturers to find the root causes of a problem and improve the manufacturing process by influencing correction strategies \cite{112-ref2, 112-ref3}. \par
This work focuses on a real-world textiles manufacturing environment. The company manufactures woollen yarns and carpets from sheep wool via chemical and mechanical processes from multiple plants. The data used is from throughout the product manufacturing process and ends with the quality "grade" that is manually given to the finished product upon final inspection.\par
The complexity of this application lies in the use of a natural fibre (sheep wool). Sheep wool has variable parameters influenced by growing conditions, season of harvesting, breed, and the parts of the sheep the wool was shorn from, among other factors. Additionally, products can be made from multiple components for design and other reasons. This multi-component manufacturing adds to the complexity of the manufacturing process. \par

\begin{figure}[h] 
  \centering
  \includegraphics[scale=0.3]{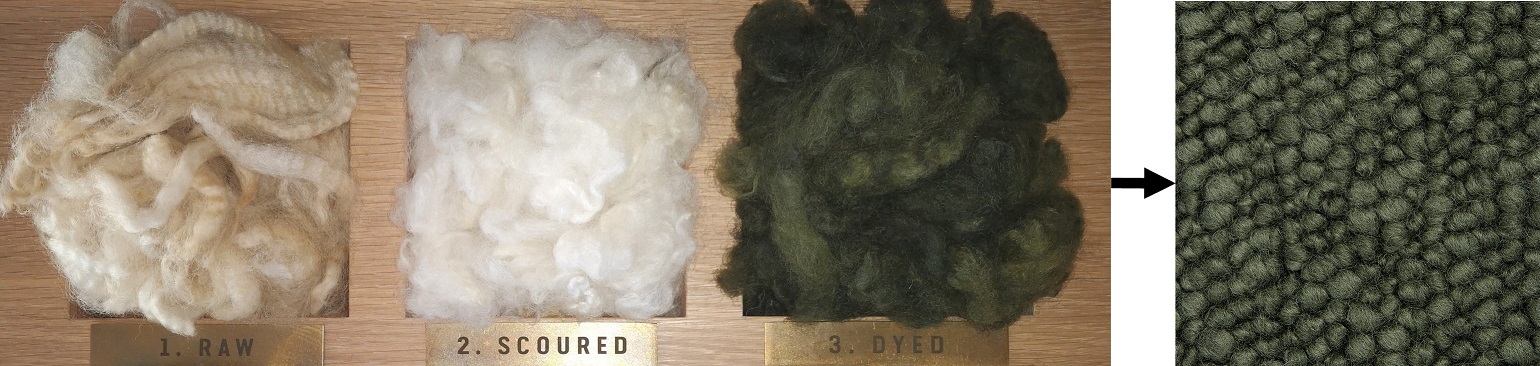}
  \caption{Select states of a woollen yarn and carpet product manufactured by Bremworth. From left to right: raw wool, scoured wool, dyed wool, the final carpet.}
  \label{wool2carpet}
\end{figure}

As seen in Figure \ref{wool2carpet}, the colour of a woollen textile product can change throughout its manufacturing process. The colour of raw wool is quantified by values that describe the brightness and yellowness of the wool \cite{wonz-1}. After scouring (cleaning) the raw wool, a core sample is taken and measurements of many key material attributes, including colour, are reported. The base colour of the wool and colour physics are referenced when calculating dye formulations. Finally, the many processing steps to achieve texture, lustre, strength, and feel throughout the yarn and carpet manufacturing processes can also affect perceived colour. \par
The work presented in this paper aimed to develop an RCA system to determine what parameter values cause a finished textile product to be outside of acceptable tolerance for colour. In other words, why does the final colour of a carpet product not match the standard? The paper also aims to present where expert knowledge is beneficial and where machine learning can be used to expand further on this knowledge. \par
The rest of this paper is organised as follows: Section \ref{related work} presents the problem domain and briefly discusses existing research and literature in the related area. The proposed methodology and solution are then outlined in Section \ref{methodology}. Subsequently, the training and evaluation results are presented and discussed in Section \ref{results}. Finally, the conclusions and areas for future work are provided in Section \ref{conclusions}.\par

\section{Related Work}
\label{related work}

In this section, some background information is presented regarding the central ideas of this paper. In Section \ref{acc_v_exp}, the trade off between model accuracy and model explainability is discussed, in Section \ref{explainable_ml}, types of explainability techniques are outlined, and finally, in Section \ref{explainable_ml_manuf}, RCA solutions developed in literature for manufacturing applications using explainable ML are described.\par

\subsection{Accuracy vs Explainability}
\label{acc_v_exp}

Model Explainability refers to models that have mechanisms in place that can justify their outputs and decisions, enabling trust and understanding between humans and an automated RCA and decision-making system \cite{89-devi}.\par
In \cite{95-shah}, the authors note that while their deep neural network (DNN) achieved better prediction performance than other evaluated methods, there were several limitations associated with deep learning (DL) techniques for industrial applications. Complex network structure makes model explainability extremely difficult and the large number of hyperparameters are difficult to tune intuitively.\par
In \cite{86-ayvaz}, the authors approached detecting errors in production systems as an anomaly detection problem. However, it was difficult to explain the results and it was not clear why the algorithm had marked a particular cycle as abnormal.
The authors concluded that it is very important to use expert knowledge, for validating and selecting data, and explainable models, for adoption and successful integration. \par
In \cite{89-devi}, the authors not only evaluated their proposed solution for accuracy but also for model explainability. Shap values (SHapley Additive exPlanations) were used which represent the contribution of a particular data point in predicting the output, which helped explain the importance of any given feature. 

\par
\subsection{Explainable ML Techniques}
\label{explainable_ml}
\begin{figure}[h] 
  \centering
  \includegraphics[scale=0.25]{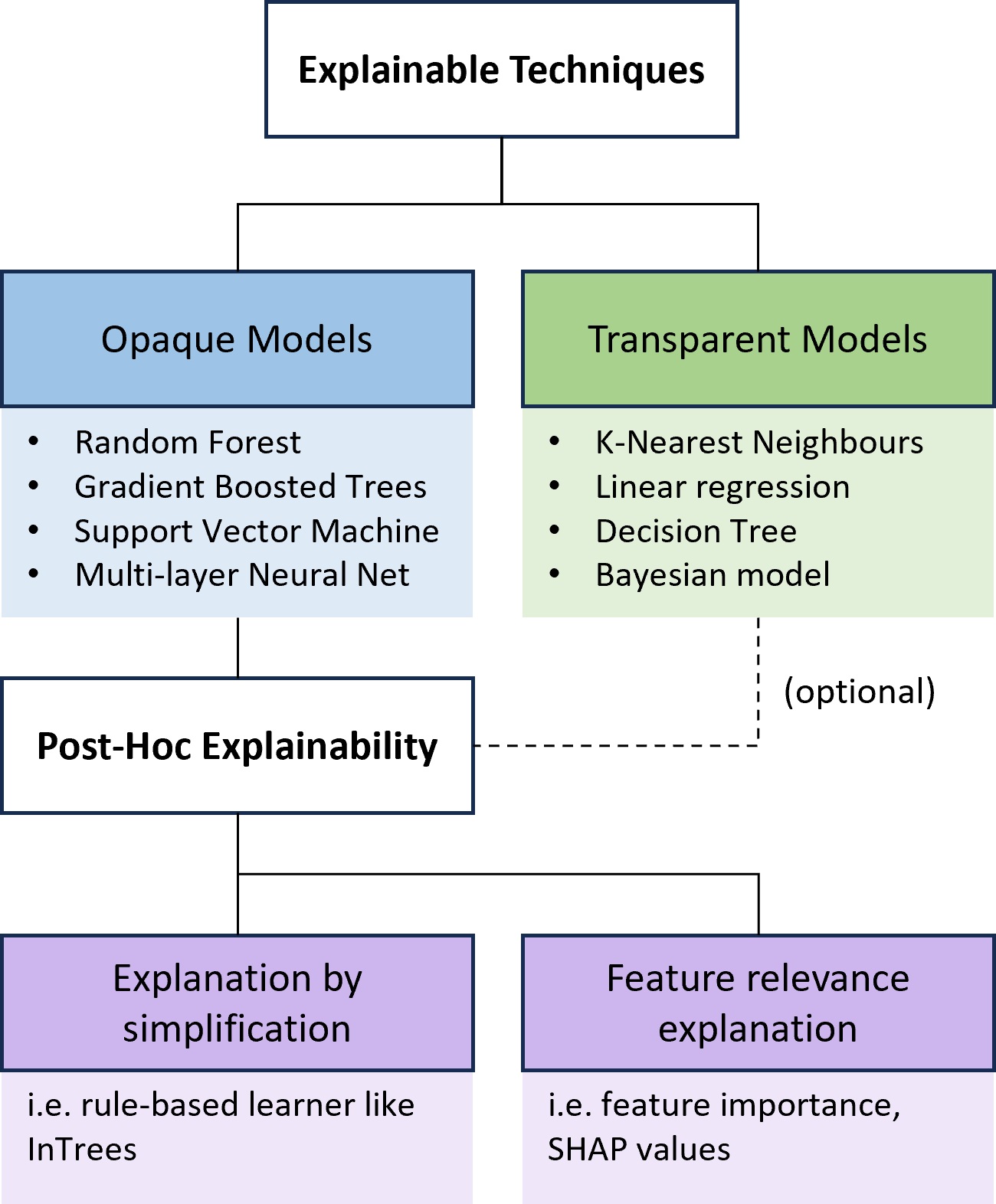}
  \caption{Examples of explainable techniques and their respective interpretation mechanisms.}
  \label{explainable breakdown}
\end{figure}

The authors of \cite{115:XAI} provide a breakdown of explainable machine learning (ML) techniques and their respective advantages and disadvantages based on the application.  In \cite{115:XAI}, a taxonomy of explainability approaches was described, with a very simplified diagram of the techniques relevant to this work shown in Figure \ref{explainable breakdown}. 

\subsubsection{Explainable Models}
With respect to explainable models, these can be split into transparent models and opaque (black box) models. The former are typically simpler and easier to understand than black box models (see Figure \ref{explainable breakdown} for examples). However, due to their simplicity, these models are less capable of representing a larger dataset featuring complex interactions. Therefore, for higher accuracy, more complex and expressive models are required.\par
Black box models include neural networks or complex ensembles of simpler models (for example ensembles of decision trees). However, the architecture of these models make it difficult to trace model outputs/predictions to specific causative input features and the relationships between features. 

\subsubsection{Post-Hoc Techniques}

These techniques take place after a model has been trained and seek to understand how input features contributed to a model's output, thereby increasing a model's transparency. \par 
In the case of interpreting tree ensembles (ensembles of simple models), rule-based learners or tree interpreters can be used to convert the complex model into something more understandable \cite{114:XAI}. The rule-based learner method extracts feature combinations and value thresholds from averaging over the variance of the simple models, whereas the Tree interpreter method decomposes every prediction result to a sum of feature contributions and bias. \par
Another post-hoc technique is using SHAP values \cite{114-40}, which can be used to explain the output of any ML model (model agnostic). The SHAP value  measures the impact of having a certain value for a given feature in comparison to the prediction. Each data instance (observation) gets assigned its own SHAP value set which then aids in explaining the prediction result for an instance and what the specific contributions of its predictors are. 

\par
\subsection{Explainable ML in Manufacturing}
\label{explainable_ml_manuf}

Common RCA methods used in the manufacturing domain include Bayesian networks, rule mining methods, and tree-based networks. There are a few areas within the manufacturing domain where ML can be applied including process control, predictive maintenance, and quality prediction. This section references studies focused on the application of ML for quality estimation and fault prediction. \par
\subsubsection{Bayesian Methods}
Bayesian networks are probabilistic graphical models, whereby a graph expresses the conditional dependency structure between random variables. 
In \cite{109:lokrantz18}, the authors used Bayesian networks to model the causal relationships between manufacturing stages using expert knowledge and evaluated them on a simulated process. The use case in \cite{109:lokrantz18} consisted of six process steps and four final quality labels that each product instance was classified against. \par
Initially, Structure Learning was used in \cite{109:lokrantz18} to automatically learn dependencies between variables, however the performance results were far below the results yielded by a model with full structure provided. The authors concluded how important accurate and in-depth expert knowledge is to the performance of the networks. A challenge in industrial applications however is that the causal effects the random variables have on the outcomes are not always clear or known.\par
\subsubsection{Rule Mining Methods}
Association Rule Mining (ARM) methods mine for frequent patterns between features and are unsupervised learning techniques \cite{111:ong15}. Unlike in Bayesian Networks, the relationships between features are based on co-occurrence instead of inherent properties extracted from expert knowledge.\par
The authors of \cite{64-Kao} evaluated their classification method on the SECOM dataset \cite{secom}. Applying the Apriori ARM method, the authors were able to extract 483 valid rules to classify each sample. To better avoid bias towards the majority class, the authors of \cite{111:ong15} developed a weighted ARM (WARM) solution whereby each feature was assigned a weight to reflect its importance. Using principal component analysis (PCA) to automatically determine feature importances, the developed solution successfully determined failure based on categorical process data (operator ID, product type etc.). \par 
\subsubsection{Tree-based Methods}
Tree-based methods have flowchart-like structures where each node represents a variable condition, each branch represents the outcome from the origin node, and the final "leaves" represent the ultimate class label or value. Decision Trees (DT) are the most rudimentary of these methods and must be manually constructed to represent a production process (for example). The Random Forest (RF) method is a bagging ensemble of randomised DTs that have a weighted combined output. The Gradient Boosted Tree (GBT) method is a boosting ensemble of randomised DTs that are trained iteratively in order to minimise a loss function.\par
With respect to real-time analysis, DT models tend to produce results much faster than other popular classification algorithms \cite{94-cakir}. However, the quality and accuracy of the construction of a DT influences the network's ability to obtain high accuracy \cite{94-cakir}.\par
In \cite{86-ayvaz}, RF and GBT methods outperformed other evaluated algorithms, including a neural network (NN) and support vector regressor (SVR). In an empirical study of supervised learning techniques evaluated over many different performance metrics, GBTs tended to outperform the RFs \cite{90-17}. The Bosch production line performance Kaggle challenge (2016), sought solutions for effective production line monitoring. Many published solutions (1st, 3rd, and 7th) used GBT-based methods for their solutions \cite{91-fahle}.\par
The approach described in \cite{65-Mangal} used an Extreme Gradient Boosting (XGB) classifier, an ensemble of weak trees, to predict class probabilities of pass or fail. The authors developed the solution in response to the Bosch production line performance Kaggle challenge (2016). To achieve the best predictions, the authors of \cite{65-Mangal} suggested that dedicated models need to be fit to each product category.\par
The system proposed in \cite{63-Syafrudin} aimed to tackle outlier detection in sensor data from an automotive manufacturing assembly line in Korea. The solution utilised Density-Based Spatial Clustering of Applications with Noise (DBSCAN) and fault prediction via Random Forest (RF) classification. Each decision tree inside the RF generated a prediction output and a majority vote was applied to obtain the final output to determine whether the manufactured product would pass or fail.\par
The SECOM dataset (SEmiCOnductor Manufacturing) \cite{secom} was used in \cite{90-moldovan} to evaluate the classification capabilities of the Random Forrest (RF) and Gradient Boosted Trees (GBT) algorithms. The results showed that the best model was obtained when the minority (Failure) class was under-sampled (see Section 3B), the features were specially selected, and the data was classified using the RF algorithm.\par
Both rule mining methods and tree-based methods present potential for use in industry due to their interpretable decision-making systems. The performance of these methods does not require expert knowledge beyond the data preparation phase.
It is unclear how well rule mining for associations between features, which was founded around the market basket paradigm using categorical data, can scale to high-dimensional numerical data from continuous distributions. However, rules can be mined from tree ensembles, where feature associations are directly linked to a prediction output \cite{114:XAI}. For this research, the use of tree-based methods with post-hoc rule mining will be explored to demonstrate prediction potential on a high-dimensional numerical dataset.

\section{Methodology}
\label{methodology}

The authors of \cite{112:overlap} noted that it is not easy to develop RCA solutions because, when considering such complex manufacturing processes, it is necessary to ensure that the methods used are adequate to the characteristics of the data available. In this research, we aimed to develop an RCA method for a textiles manufacturing process that is interpretable and can accurately determine, from early-stage process data, when a product instance might deviate from the standard.\par
The following sections describe how existing techniques were bench-marked to compare their behaviour in this domain and how the solution presented in this paper was developed.

\begin{figure}[ht] 
  \centering
  \includegraphics[scale=0.3]{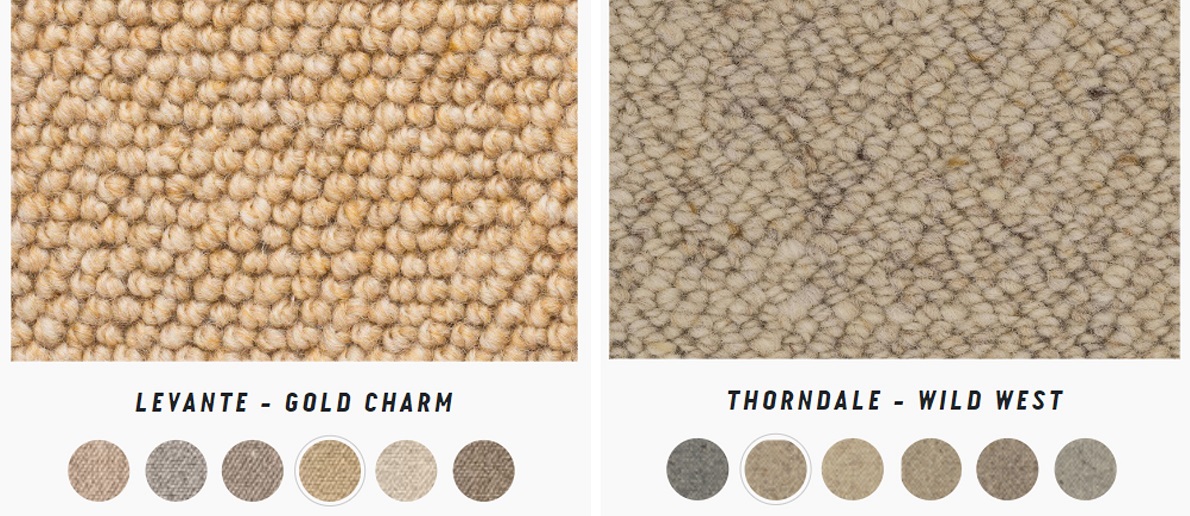}
  \caption{Examples of the two products that were investigated in this work and their respective colour shades.}
  \label{product examples}
\end{figure}

\subsection{Dataset}
\label{dataset}

The dataset used in this study was obtained from a New Zealand manufacturer of woollen carpets and rugs. The dataset was selected from a period covering three years. The authors of \cite{65-Mangal} suggested that, in order to achieve the best predictions, dedicated models should be fit to each product category. As such, two products of similar design were investigated in this research. These can be seen in Figure \ref{product examples}. \par
Because process data is typically highly imbalanced between 1) the success and failure cases and 2) the different failure classes themselves, the authors of \cite{86-ayvaz} omitted rare failure types in their dataset and focused on two failure types only. Likewise, for this work, the most common defect category was selected with other defective instances discarded from the dataset. In practice, these defect categories are assigned at the final inspection stage of the finished product and, in this case, represent the miscolouration of the product from the standard.\par
In total, after preprocessing (discussed in the Section \ref{data_processing}), there were 300 instances of the selected products. 210 of these were compliant (Label = 0) and the remainder were marked as off-colour (Label = 1).\par

\begin{figure}[ht] 
  \centering
  \includegraphics[scale=0.17]{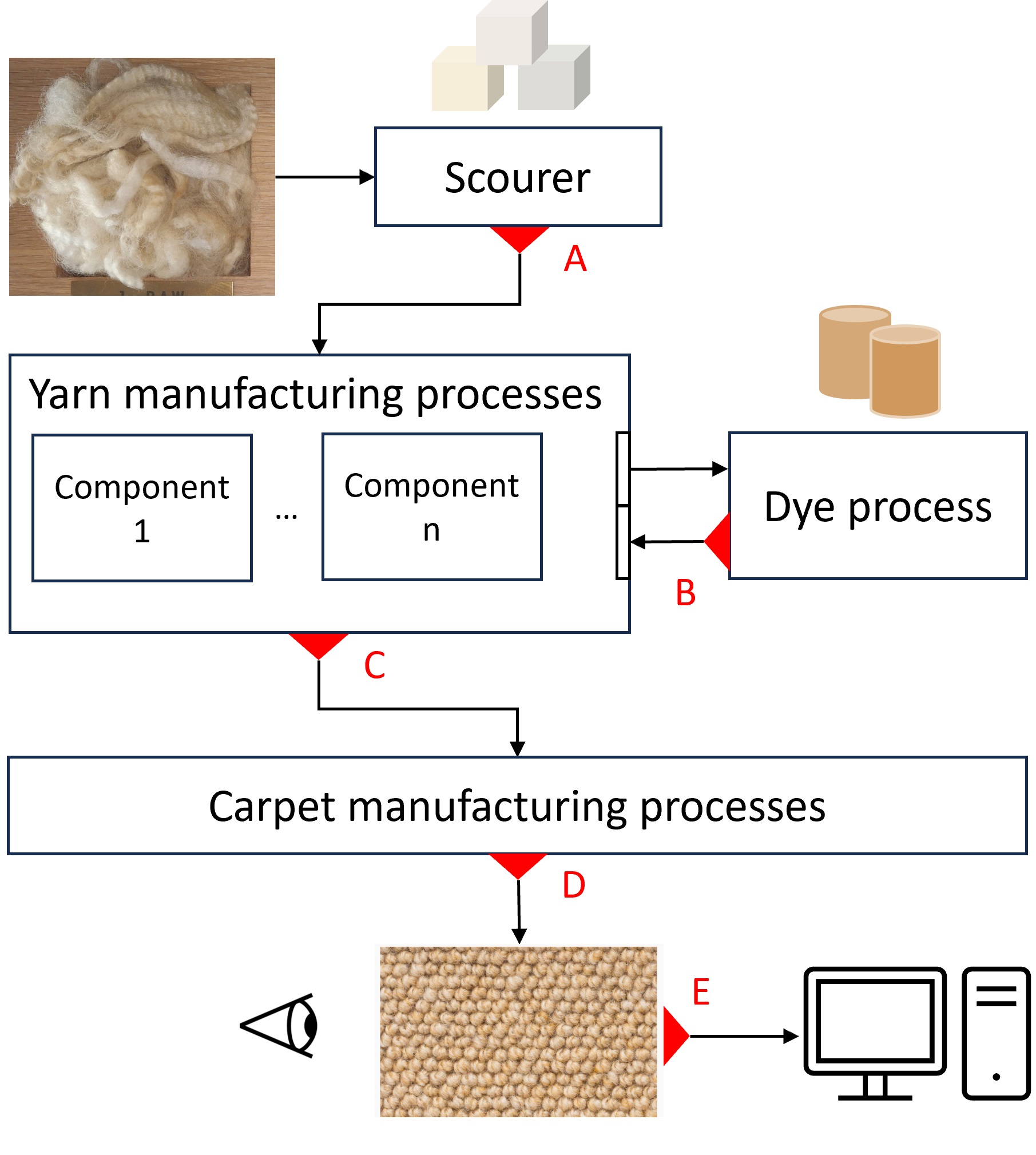}
  \caption{A high-level overview of the wool-to-yarn-to-carpet manufacturing process, with data point locations indicated in red.}
  \label{process}
\end{figure}

With respect to the features of the dataset, following the feature engineering steps discussed in Section \ref{data_processing}, the complete dataset has 372 columns. The features include information about:
\begin{itemize}
\item Raw material variables (Data Point A in Figure \ref{process}): these features relate to the properties of the wool after scouring and include, but are not limited to, the base and as-is colours (measured as X,Y,Z tristimulus values), mean fibre length and diameter (in mm), and percentages of vegetable matter content, grease content, and moisture content.
\item Manufacturing variables (data points B, C, and D in Figure \ref{process}): the product specifications are captured amongst these features and include all the colourants and their respective quantities (in g or mL), and any additional additives that may be added for process efficiency or product purposes.
\item Quality control variable (Data Point E in Figure \ref{process}): finally, a single target variable was used which is the colour grade manually given to the finished carpet upon final visual inspection.
\end{itemize}

\subsection{Data Preprocessing}
\label{data_processing}

A single carpet batch is made up of multiple manufactured components, each represented as an instance within the dataset (sub-instances). In combining these sub-instances for a single representation of the completed batch/instance, the features of each sub-instance were weighted by their respective contributing quantities. Finally, all features were averaged over the sub-instances for a single batch/instance. Additional statistical features were captured and appended to the dataset as new features, including the minimum and maximum of each sub-instance of a batch/instance.\par
Common challenges of raw process data include missing data, diverse value ranges, and imbalanced data. \par
To address missing data, similar to \cite{86-ayvaz, 90-moldovan}, any feature column for which more than 55-60\% of the data was missing, was removed. Where missing data was more isolated, undefined values were filled in with the median values of the respective columns. Conversely, outliers were replaced with feature column minimums or maximums, whichever was appropriate. \par
With respect to the diverse value ranges across features, the authors of \cite{86-ayvaz, 90-moldovan} noted that it is important to normalise features that are not on the same scale when training classification models. In this work, all values were normalised against the corresponding product quantities (kg) and then each feature column was normalised via min-max normalisation such that all values were within the range of [0,1]. \par

\subsection{Feature Engineering} 
\label{feature_engineering}

Feature engineering, or feature fusion, is where additional features are added to a dataset, to bring in new information that may benefit the classification process. These new features can either be statistical and/or knowledge domain-based. As will be discussed in Section \ref{results}, the new features that were added to the dataset for this work had a major influence on the performance of the evaluated classifiers. This seems to fall in line with the findings of \cite{65-Mangal} where the authors noted that additional time features, including durations and mean time-to-failures, improved the performance of their developed solution.\par
For this work, a choice was made to separate features of the components that were coloured during the manufacturing process from the components that were not. This resulted in a concatenated dataset, whereby the features were duplicated but the respective amounts for each instance were weighted according to the proportions of the coloured and uncoloured component quantities. From here, following expert advice and formulae from literature \cite{woolcolour1, woolcolour2, woolcolour3} of potentially relevant variables, new features were calculated from existing features and added to the dataset, including:
\begin{enumerate}
    \item hue of components
    \item depth of colour of components 
    \item differences between colour values measured under different conditions or at different stages of the process
\end{enumerate}

Throughout the analysis stage of this work, an iterative approach was taken on the treatment of missing data and feature engineering to improve model performance to a level that was useful for prediction. \par

\subsection{Feature Selection} 
\label{feature_selection}

Feature selection is a technique used to identify the top relevant features from a larger set of features in a dataset. By reducing the number of features used in an ML algorithm, issues such as overfitting, multicollinearity, computational complexity, and reduced interpretability can be avoided \cite{86-ayvaz}. In this work, three feature selection methods were evaluated.

\subsubsection{SelectKBest (Chi-Squared)}

A statistical test for measuring the association between two nominal factors is the chi-square independence test. This test is performed between each feature in a dataset and the respective target label. Using the chi-square test, the probabilities that two variables  are independent are calculated. The aim of the method is to determine features that are dependent on the target and thus it is typical to select features that have probabilities less than 0.05. Finally, the SelectKBest function uses the scores from the chi-square test and selects the top 'K' number of features for further analysis, where K is manually set.

\subsubsection{Pearson Correlation Coefficient}

\begin{figure}[h] 
  \centering
  \includegraphics[scale=0.3]{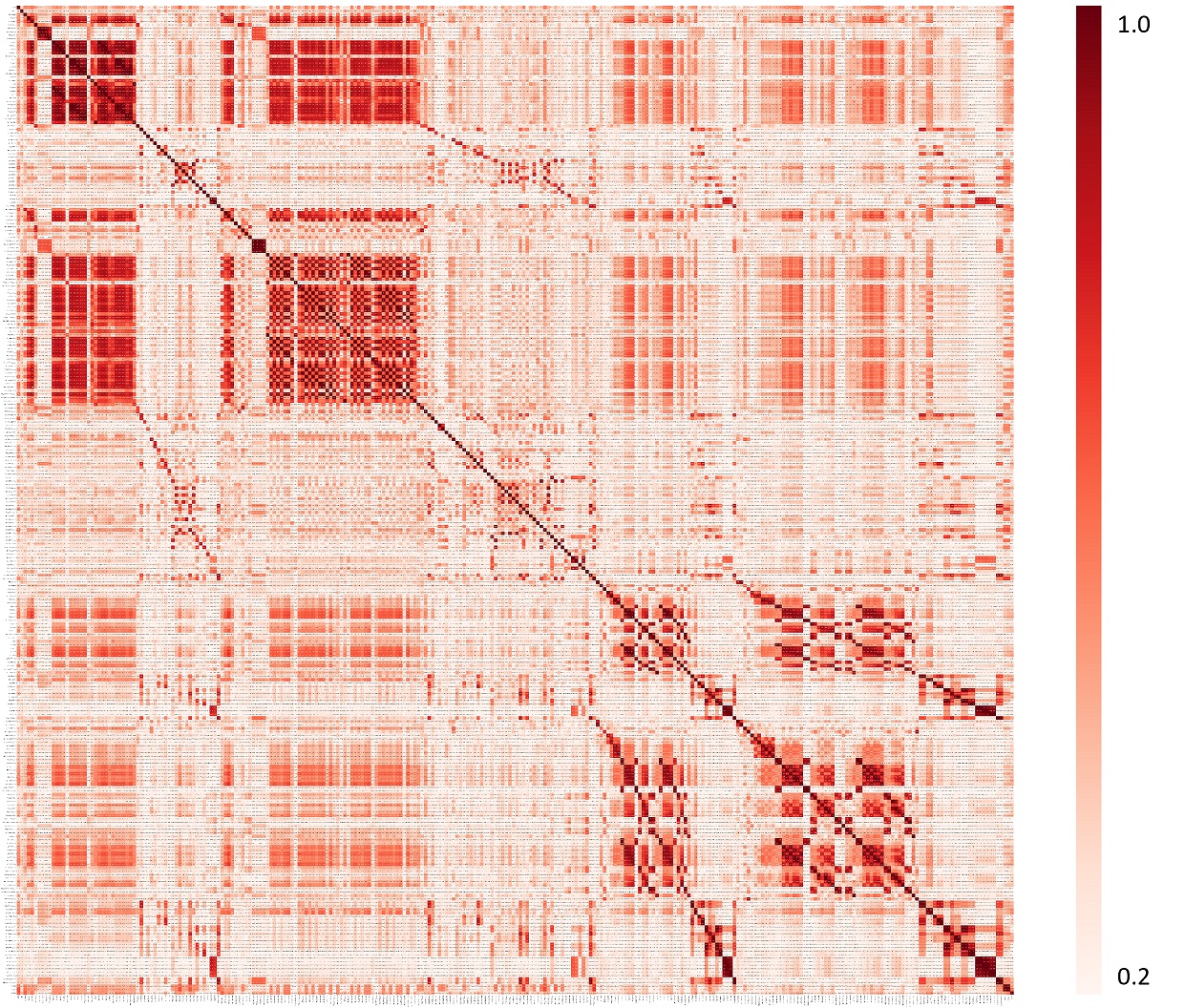}
  \caption{Correlation heatmap using the Pearson Correlation Coefficient}
  \label{correlation heatmap}
\end{figure}

The pearson correlation coefficient (PCC) values capture two separate correlation relations between each feature and the target label. Firstly, the correlations of each feature with the target and secondly, the correlations between the features themselves. Features that are estimated as being highly correlated with the target should be selected while one feature from a highly correlated feature pair can be removed (seen as redundant information). Two threshold values are manually set to determine the cutoff points for what features are considered to be "highly correlated". As can be seen in Figure \ref{correlation heatmap}, features exist within the dataset that have high correlation with other features and conversely, there are features that have little correlation with the target label.

\renewcommand{\arraystretch}{1.25}
\begin{table*}[htb!]
\small 
\centering
\caption{Results from testing three different tree-based classifiers on three different feature reduction techniques. Bold values indicate the best results for the respective dataset, yellow values indicate the best results for the respective model.}
\label{table_results}

\begin{tabular}{l|ccc|ccc|ccc|ccc|}
\cline{2-13}
\textbf{} & \multicolumn{3}{c|}{\textbf{All Features}} & \multicolumn{3}{c|}{\textbf{SelectKBest (chi-square)}} & \multicolumn{3}{c|}{\textbf{Pearson Coefficient}} & \multicolumn{3}{c|}{\textbf{Boruta Algorithm}} \\ \hline
\multicolumn{1}{|l|}{\textbf{Models}} & \multicolumn{1}{c|}{Precision} & \multicolumn{1}{c|}{Recall}        & F1 & \multicolumn{1}{c|}{Precision} & \multicolumn{1}{c|}{Recall} & F1 & \multicolumn{1}{c|}{Precision} & \multicolumn{1}{c|}{Recall} & F1 & \multicolumn{1}{c|}{Precision} & \multicolumn{1}{c|}{Recall} & F1 \\ \hline
\multicolumn{1}{|l|}{Decision Tree} & \multicolumn{1}{c|}{\cellcolor[HTML]{FFFE65}{\textbf{0.74}}} & \multicolumn{1}{c|}{\textbf{0.68}} & {\cellcolor[HTML]{FFFE65}\textbf{0.71}} & \multicolumn{1}{c|}{0.69} & \multicolumn{1}{c|}{0.68} & 0.68 & \multicolumn{1}{c|}{0.67} & \multicolumn{1}{c|}{\textbf{0.68}} & 0.67 & \multicolumn{1}{c|}{0.7}& \multicolumn{1}{c|}{\cellcolor[HTML]{FFFE65}0.7} & 0.7 \\ \hline
\multicolumn{1}{|l|}{Random   Forest} & \multicolumn{1}{c|}{0.73} & \multicolumn{1}{c|}{0.63} & 0.67 & \multicolumn{1}{c|}{0.76} & \multicolumn{1}{c|}{0.69} & 0.72 & \multicolumn{1}{c|}{\textbf{0.76}} & \multicolumn{1}{c|}{0.65} & \textbf{0.69} & \multicolumn{1}{c|}{\cellcolor[HTML]{FFFE65}\textbf{0.81}} & \multicolumn{1}{c|}{\cellcolor[HTML]{FFFE65}\textbf{0.71}} & {\cellcolor[HTML]{FFFE65}\textbf{0.76}} \\ \hline
\multicolumn{1}{|l|}{XGBoost} & \multicolumn{1}{c|}{0.73} & \multicolumn{1}{c|}{\textbf{0.68}} & 0.7 & \multicolumn{1}{c|}{\cellcolor[HTML]{FFFE65}{\textbf{0.77}}} & \multicolumn{1}{c|}{\cellcolor[HTML]{FFFE65}{\textbf{0.72}}} & {\cellcolor[HTML]{FFFE65}\textbf{0.74}} & \multicolumn{1}{c|}{0.73} & \multicolumn{1}{c|}{0.62} & 0.66 & \multicolumn{1}{c|}{\cellcolor[HTML]{FFFE65}0.77} & \multicolumn{1}{c|}{0.7} & 0.73 \\ \hline
\end{tabular}

\end{table*}

\subsubsection{Boruta Algorithm}

The Boruta algorithm is a wrapper algorithm around an RF classifier. Firstly, the algorithm adds randomness to a dataset by creating shuffled copies of all features (called shadow features). Then, it trains an RF classifier on this new dataset and calculates the importances of each feature. During training, the algorithm checks every iteration whether a real feature has a higher importance than the best of the shadow features and, if so, adds to an importance counter for the respective feature. Finally, after a specified number of iterations, a threshold is automatically determined from a binomial distribution based on the number of iterations. Any features with an importance counter above this are deemed important and kept for further analysis.\par
The authors of \cite{90-moldovan} used the Boruta algorithm and others for the selection of the most relevant features in their semiconductor manufacturing dataset. During evaluation, along with other techniques, the Boruta algorithm contributed to the best performing classifier \cite{90-moldovan}.

\subsection{Prediction Models}
\label{models}

In quality prediction, the class of an instance (a finished product) is predicted as either normal or abnormal. For this work, three different tree-based classification algorithms were evaluated: a Decision Tree (DT) and two ensemble methods, Random Forest (RF) and XGBoost (XGB). These algorithms were selected because (1) mechanisms exist for supporting interpretability, and (2) the algorithms have a demonstrated track record from literature for high prediction accuracy on big data.\par 
As previously discussed, a single DT may not be enough to yield optimal performance. Thus, several DTs can be combined as an ensemble to improve model performance. However, model complexity of ensembles increases with the number of decision trees and, as a result, model interpretability decreases. TE2Rules \cite{te2rules} is a technique to explain complex tree ensembles such as XGB and RF, trained on a binary classification task, using a rule list. The extracted rule list captures the necessary and sufficient conditions for classification by the tree ensemble. The algorithm used by TE2Rules is based on Apriori Rule Mining. \par
For hyperparameter tuning, random search with 5-fold cross validation technique was used. For the XGB model, the optimal hyperparameters were found to be a maximum depth of 8, 121 tree estimators, and the Gini Impurity criterion. Similarly, the optimal parameters for the RF model were found to be a maximum depth of 75, 92 tree estimators, and the Gini Impurity criterion. \par
The classification models and dataset processing steps were implemented and evaluated using the Pandas and scikit-learn libraries in Python. Other specific libraries used include xgboost (boosting classification model), te2rules (tree ensemble explainer), and shap (feature importances). 10-fold stratified cross-validation was used to split the dataset for model training and evaluation. The results were averaged over these 10 seeds to provide a measure of how well the models generalise to new data.

\section{Results and Discussion}
\label{results}

From the results in Table \ref{table_results}, it can be seen that all models achieve similar performance when presented with the complete dataset. Across the feature reduction techniques, it is notable that, unlike the Decision Tree, the ensemble methods see a boost in performance in one or more of the models using a reduced dataset. Overall, the Random Forest method yielded the best results when combined with the Boruta algorithm. \par
When it comes to the application for this work, incorrectly predicting an instance as of acceptable colour/normal is not as bad as incorrectly predicting an instance as unacceptable/downgraded. This is because a potential use of these classifiers is to catch issues early on in the manufacturing process such that remedial action may be taken in preventing the end product from being downgraded. Thus, proportions of actual downgrades being captured (recall) and proportions of predicted downgrades that are correct (precision) are two metrics that are of interest. The definitions and formulae for these evaluation metrics can be found in Table \ref{table_metrics}. \par

\renewcommand{\arraystretch}{1.25}
\begin{table}[h]
\footnotesize 
\caption{(a) Confusion matrix definitions, (b) Evaluation metrics for classification models}
\label{table_metrics}
\resizebox{\columnwidth}{!}{%
\begin{tabular}{llcl}
\multicolumn{4}{c}{(a)} \\ \cline{3-4} 
\multicolumn{2}{l|}{} & \multicolumn{1}{c|}{\textbf{Predicted downgrade}} & \multicolumn{1}{c|}{\textbf{Predicted normal}} \\ \hline
\multicolumn{2}{|l|}{\textbf{\begin{tabular}[c]{@{}l@{}}Ground truth\\ downgrade\end{tabular}}} & \multicolumn{1}{c|}{\begin{tabular}[c]{@{}c@{}}Correct downgrade\\ (TP: True Positive)\end{tabular}} & \multicolumn{1}{c|}{\begin{tabular}[c]{@{}c@{}}Missed downgrade\\ (FN: False Negative)\end{tabular}} \\ \hline
\multicolumn{2}{|l|}{\textbf{\begin{tabular}[c]{@{}l@{}}Ground truth\\ normal\end{tabular}}} & \multicolumn{1}{c|}{\begin{tabular}[c]{@{}c@{}}Incorrect downgrade\\ (FP: False Postive)\end{tabular}} & \multicolumn{1}{c|}{\begin{tabular}[c]{@{}c@{}}Correct normal\\ (TN: True Negative)\end{tabular}} \\ \hline
\multicolumn{4}{c}{(b)} \\ \hline
(1) & \multicolumn{2}{c}{Recall = $\frac{TP}{TP + FN}$} & \begin{tabular}[c]{@{}l@{}}Ratio of all downgraded \\ instances correctly classified\end{tabular} \\
(2) & \multicolumn{2}{c}{Precision = $\frac{TP}{TP + FP}$} & \begin{tabular}[c]{@{}l@{}}Ratio of all classified \\ downgraded instances that\\ are correct\end{tabular} \\
(3) & \multicolumn{2}{c}{FPR = $\frac{FP}{FP + TN}$} & \begin{tabular}[c]{@{}l@{}}Ratio of all normal instances\\ incorrectly classified as\\ downgrades\end{tabular} \\
(4) & \multicolumn{2}{c}{F1 = $\frac{2 . Recall . Precision}{Recall + Precision}$} & \begin{tabular}[c]{@{}l@{}}Harmonic mean between\\ Recall and Precision\end{tabular} \\ \hline
\end{tabular}%
}
\end{table}

It can be seen in Table \ref{table_results} that the RF model did the best with respect to yielding the highest Precision values. However, for all feature selection techniques apart from Boruta, it would seem that this higher precision was at the expense of recall, whereby the percentage of correctly classified downgrades versus actual downgrades is captured. \par
Finally, it is notable that there is a shortfall in performance, with most results being below 80\%. This performance gap could be partially attributed to missing features in the dataset of manufacturing variables that may have an effect on the final outcome of the batch. Training the models on a larger dataset, with a more comprehensive range of acceptable instances and downgraded instances, could improve prediction performance. However, though small, the dataset used for this work yielded results that are still statistically meaningful and demonstrate potential for industry use to influence manufacturing decisions. \par
With respect to model interpretability and performance trade-offs, it can also be seen from Table \ref{table_results} that the main distinction in performance between the Decision Tree (DT) method and the tree ensemble methods is in the model's precision. Across the different feature selection techniques, the negative difference between the DT and the respective best performing model is between 6 and 11\%. As discussed previously, because this metric is of particular importance for this application, this difference in performance could be seen as sufficient evidence for selecting an opaque model over a transparent model. \par
\begin{figure}[h] 
  \centering
  \includegraphics[scale=0.23]{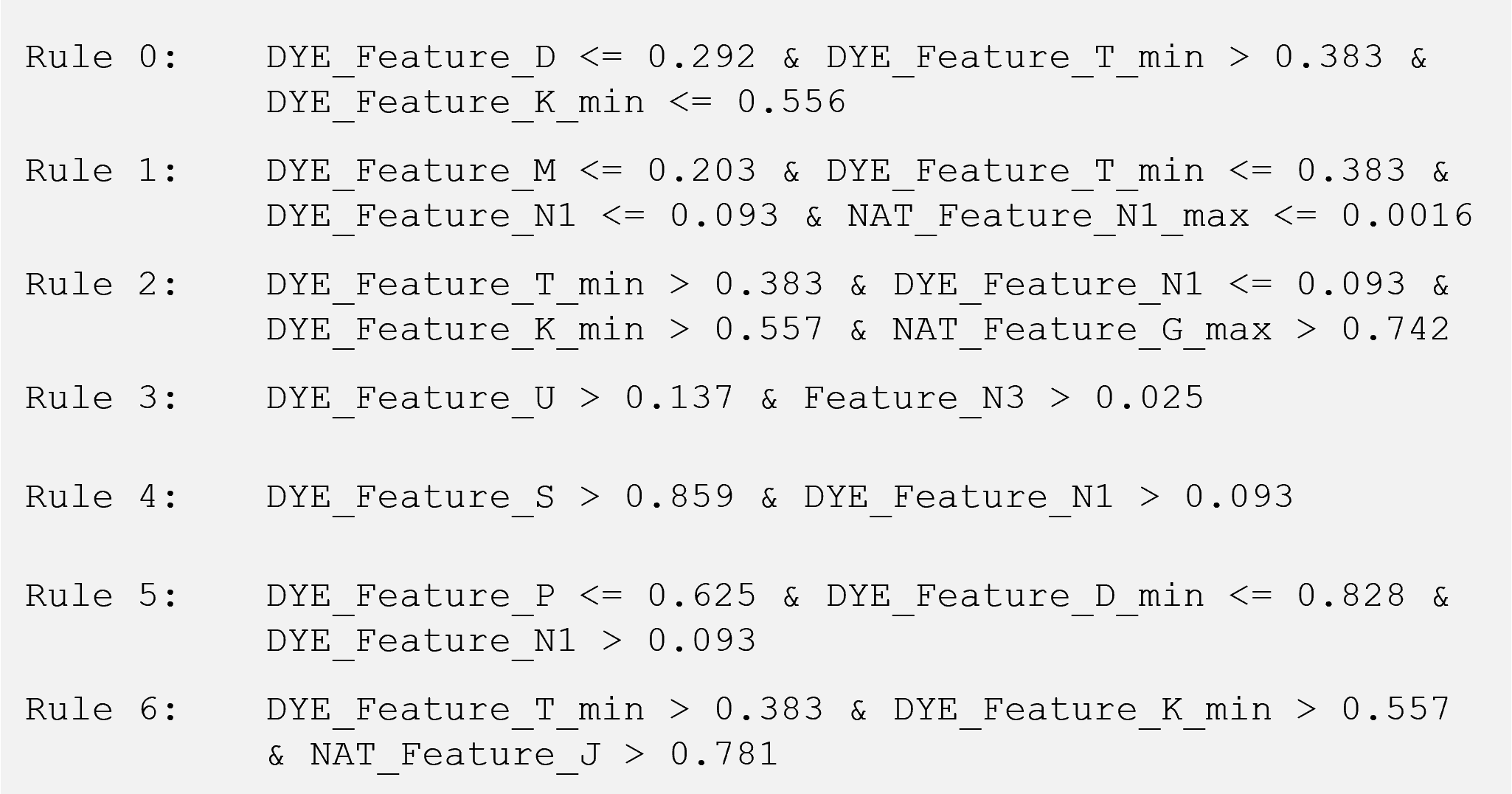}
  \caption{Examples of rules extracted from the RF model w/ Boruta feature selection using the TE2Rules python package \cite{te2rules}}
  \label{feature_rules}
\end{figure}

This leads into the post-hoc explainability technique to extract rules from the trained tree ensemble methods. In this case, rules were extracted using the TE2Rules method discussed in Section \ref{models}. An example of the rules extracted can be seen in Figure \ref{feature_rules}. 

As can be seen in Figure \ref{feature_rules}, several features that were singled out for prediction rules were additional statistical features (minimum and maximum feature columns) and new ones calculated from original features and relevant formulae (Feature\_N1). From the 372 features that made up the complete dataset, this demonstrates the influence that additional features can have on the ultimate model performance. The iterative data preprocessing stage also reinforced the importance of understanding the problem domain and relevant data features. \par
Additionally, it was found that the rules had naturally segmented the dataset based on discriminative product features. With this knowledge, the technical experts could analyse the extracted rules on a product feature basis. Certain rules proved insightful for predicting failure for certain product features, ultimately informing more targeted manufacturing decisions, based on insights that had been generated from the full dataset.

\section{Conclusions}
\label{conclusions}

In this paper, the use of explainable ML techniques was investigated for the purpose of predicting colour quality in a complex textiles manufacturing process. 
Manufacturing woollen yarns and carpets is a complex process due to the variability of working with a natural material such as wool. The dataset used was provided by a New Zealand-based manufacturer of woollen carpets and rugs and had a high-dimensional feature space and class imbalance. The data was classified using three tree-based methods, a Decision Tree (DT), a Random Forest (RF) ensemble model, and the XGBoost (XGB) ensemble model. Different feature selection methods were also investigated and included SelectKBest, Pearson Correlation Coefficient, and Boruta methods. \par
Ultimately, the explainable techniques demonstrated promise for root cause analysis in this use case. From evaluation, the combination of techniques that yielded the best results was the RF classification model and the Boruta feature selection algorithm. When paired with a rule-learner algorithm such as TE2Rules, an opaque model (RF) with post-hoc explainability techniques yielded better results than a transparent model (DT). Finally, this work highlighted the importance of additional calculated features, based on expert knowledge at the data preprocessing stage, as a way to improve model performance. \par
Future work will look into expanding the prediction capability to other products and downgrade categories. Additionally, an important next step will be to add a mechanism that will be able to suggest correction values with respect to the features that have been found to exceed, or fall short of, the rule thresholds. \par 


\section*{ACKNOWLEDGMENT}

This project was funded by Bremworth Carpets and Rugs, New Zealand.  Bremworth acknowledges the Co-Funding of this project by the Ministry for Primary Industries’ Sustainable Food and Fibre Futures Fund.

\bibliography{bibFile}

\begin{thebibliography}{10}
\providecommand{\url}[1]{#1}
\csname url@rmstyle\endcsname
\providecommand{\newblock}{\relax}
\providecommand{\bibinfo}[2]{#2}
\providecommand\BIBentrySTDinterwordspacing{\spaceskip=0pt\relax}
\providecommand\BIBentryALTinterwordstretchfactor{4}
\providecommand\BIBentryALTinterwordspacing{\spaceskip=\fontdimen2\font plus
\BIBentryALTinterwordstretchfactor\fontdimen3\font minus \fontdimen4\font\relax}
\providecommand\BIBforeignlanguage[2]{{%
\expandafter\ifx\csname l@#1\endcsname\relax
\typeout{** WARNING: IEEEtran.bst: No hyphenation pattern has been}%
\typeout{** loaded for the language `#1'. Using the pattern for}%
\typeout{** the default language instead.}%
\else
\language=\csname l@#1\endcsname
\fi
#2}}

\bibitem{112-ref1}
Z.~Li and S.~Zhou, ``Robust method of multiple variation sources identification in manufacturing processes for quality improvement,'' 2006.

\bibitem{112-ref4}
P.-L. Ong, Y.-H. Choo, and A.~K. Muda, ``A manufacturing failure root cause analysis in imbalance data set using pca weighted association rule mining,'' \emph{Jurnal Teknologi}, vol.~77, no.~18, pp. 103--111, 2015.

\bibitem{112-ref2}
S.~Sahoo, ``Big data analytics in manufacturing: a bibliometric analysis of research in the field of business management,'' \emph{International Journal of Production Research}, vol.~60, no.~22, pp. 6793--6821, 2022.

\bibitem{112-ref3}
H.~Tarakci, ``Two types of learning effects on maintenance activities,'' \emph{International Journal of Production Research}, vol.~54, no.~6, pp. 1721--1734, 2016.

\bibitem{wonz-1}
``Selection and blending of wool for carpet manufacture,'' \emph{Carpet Technical Information}, 2002.

\bibitem{89-devi}
T.~K. Devi, E.~Priyanka, and P.~Sakthivel, ``Paper quality enhancement and model prediction using machine learning techniques,'' \emph{Results in Engineering}, vol.~17, p. 100950, 2023.

\bibitem{95-shah}
D.~Shah, J.~Wang, and Q.~P. He, ``Feature engineering in big data analytics for iot-enabled smart manufacturing--comparison between deep learning and statistical learning,'' \emph{Computers \& Chemical Engineering}, vol. 141, p. 106970, 2020.

\bibitem{86-ayvaz}
S.~Ayvaz and K.~Alpay, ``Predictive maintenance system for production lines in manufacturing: A machine learning approach using iot data in real-time,'' \emph{Expert Systems with Applications}, vol. 173, p. 114598, 2021.

\bibitem{115:XAI}
V.~Belle and I.~Papantonis, ``Principles and practice of explainable machine learning,'' \emph{Frontiers in big Data}, vol.~4, p. 688969, 2021.

\bibitem{114:XAI}
R.~Dwivedi, D.~Dave, H.~Naik, S.~Singhal, R.~Omer, P.~Patel, B.~Qian, Z.~Wen, T.~Shah, G.~Morgan, \emph{et~al.}, ``Explainable ai (xai): Core ideas, techniques, and solutions,'' \emph{ACM Computing Surveys}, vol.~55, no.~9, pp. 1--33, 2023.

\bibitem{114-40}
S.~M. Lundberg and S.-I. Lee, ``A unified approach to interpreting model predictions,'' \emph{Advances in neural information processing systems}, vol.~30, 2017.

\bibitem{109:lokrantz18}
A.~Lokrantz, E.~Gustavsson, and M.~Jirstrand, ``Root cause analysis of failures and quality deviations in manufacturing using machine learning,'' \emph{Procedia Cirp}, vol.~72, pp. 1057--1062, 2018.

\bibitem{111:ong15}
P.-L. Ong, Y.-H. Choo, and A.~K. Muda, ``A manufacturing failure root cause analysis in imbalance data set using pca weighted association rule mining,'' \emph{Jurnal Teknologi}, vol.~77, no.~18, pp. 103--111, 2015.

\bibitem{64-Kao}
H.-A. Kao, Y.-S. Hsieh, C.-H. Chen, and J.~Lee, ``Quality prediction modeling for multistage manufacturing based on classification and association rule mining,'' in \emph{MATEC Web of Conferences}, vol. 123.\hskip 1em plus 0.5em minus 0.4em\relax EDP Sciences, 2017, p. 00029.

\bibitem{secom}
M.~McCann and A.~Johnston, ``{SECOM},'' UCI Machine Learning Repository, 2008, {DOI}: https://doi.org/10.24432/C54305.

\bibitem{94-cakir}
M.~Cakir, M.~A. Guvenc, and S.~Mistikoglu, ``The experimental application of popular machine learning algorithms on predictive maintenance and the design of iiot based condition monitoring system,'' \emph{Computers \& Industrial Engineering}, vol. 151, p. 106948, 2021.

\bibitem{90-17}
R.~Caruana and A.~Niculescu-Mizil, ``An empirical comparison of supervised learning algorithms,'' in \emph{Proceedings of the 23rd international conference on Machine learning}, 2006, pp. 161--168.

\bibitem{91-fahle}
S.~Fahle, C.~Prinz, and B.~Kuhlenk{\"o}tter, ``Systematic review on machine learning (ml) methods for manufacturing processes--identifying artificial intelligence (ai) methods for field application,'' \emph{Procedia CIRP}, vol.~93, pp. 413--418, 2020.

\bibitem{65-Mangal}
A.~Mangal and N.~Kumar, ``Using big data to enhance the bosch production line performance: A kaggle challenge,'' in \emph{2016 IEEE international conference on big data (big data)}.\hskip 1em plus 0.5em minus 0.4em\relax IEEE, 2016, pp. 2029--2035.

\bibitem{63-Syafrudin}
M.~Syafrudin, G.~Alfian, N.~L. Fitriyani, and J.~Rhee, ``Performance analysis of iot-based sensor, big data processing, and machine learning model for real-time monitoring system in automotive manufacturing,'' \emph{Sensors}, vol.~18, no.~9, p. 2946, 2018.

\bibitem{90-moldovan}
D.~Moldovan, T.~Cioara, I.~Anghel, and I.~Salomie, ``Machine learning for sensor-based manufacturing processes,'' in \emph{2017 13th IEEE international conference on intelligent computer communication and processing (ICCP)}.\hskip 1em plus 0.5em minus 0.4em\relax IEEE, 2017, pp. 147--154.

\bibitem{112:overlap}
E.~e~Oliveira, V.~L. Migu{\'e}is, and J.~L. Borges, ``On the influence of overlap in automatic root cause analysis in manufacturing,'' \emph{International Journal of Production Research}, vol.~60, no.~21, pp. 6491--6507, 2022.

\bibitem{woolcolour1}
J.~A. Rippon, ``The chemical and physical basis for wool dyeing,'' \emph{The coloration of wool and other keratin fibres}, pp. 43--74, 2013.

\bibitem{woolcolour2}
\BIBentryALTinterwordspacing
R.~Brady, 2001. [Online]. Available: \url{https://www.woolwise.com/wp-content/uploads/2017/05/05.1-Theory-of-Colour-Measurement-Notes.pdf}
\BIBentrySTDinterwordspacing

\bibitem{woolcolour3}
R.~Treigien{\.e}, ``The influence of physical factors on wool fibre colour changes,'' \emph{Materials Science}, vol.~16, no.~4, pp. 341--345, 2010.

\bibitem{te2rules}
G.~R. Lal, X.~Chen, and V.~Mithal, ``Te2rules: Extracting rule lists from tree ensembles,'' \emph{arXiv preprint arXiv:2206.14359}, 2022.

\end{thebibliography}
\bibliographystyle{IEEEtran.bst}


\end{document}